# Advancing Urban Renewal

An Automated Approach to Generating Historical Arcade Facades with Stable Diffusion Models

Zheyuan Kuang
Nanchang University
Jiaxin Zhang
Nanchang University
Yiying Huang
Nanchang University
Yunqin Li
Nanchang University

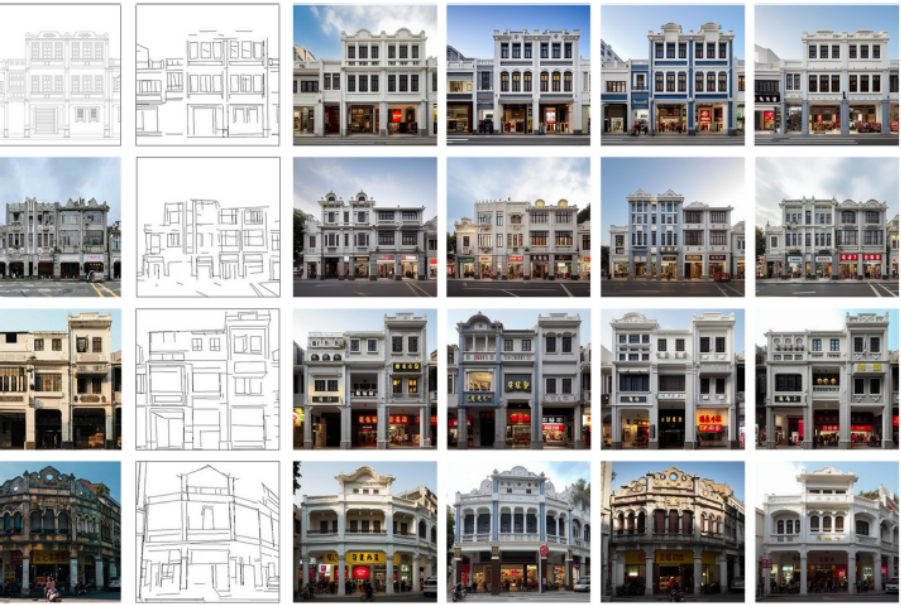
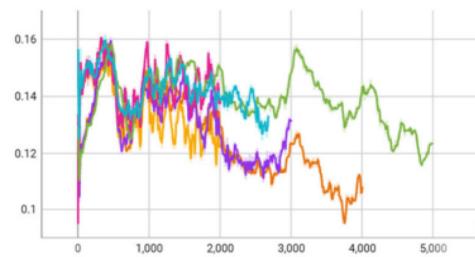
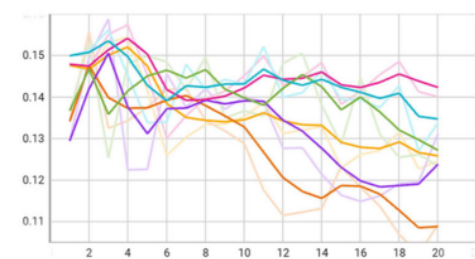

## ABSTRACT

Urban renewal and transformation processes necessitate the preservation of the historical urban fabric, particularly in districts known for their architectural and historical significance. These regions, with their diverse architectural styles, have traditionally required extensive preliminary research, often leading to subjective results. However, the advent of machine learning models has opened up new avenues for generating building facade images. Despite this, creating high-quality images for historical district renovations remains challenging, due to the complexity and diversity inherent in such districts. In response to these challenges, our study introduces a new methodology for automatically generating images of historical arcade facades, utilizing Stable Diffusion models conditioned on textual descriptions. By classifying and tagging a variety of arcade styles, we have constructed several realistic arcade facade image datasets. We trained multiple low-rank adaptation (LoRA) models to control the stylistic aspects of the generated images, supplemented by ControlNet models for improved precision and authenticity. Our approach has demonstrated high levels of precision, authenticity, and diversity in the generated images, showing promising potential for real-world urban renewal projects. This new methodology offers a more efficient and accurate alternative to conventional design processes in urban renewal, bypassing issues of unconvincing image details, lack of precision, and limited stylistic variety. Future research could focus on integrating this two-dimensional image generation with three-dimensional modeling techniques, providing a more comprehensive solution for renovating architectural facades in historical districts.

Above left: Arcade facade renewal based on prompt and ControlNet.

Above right: Training loss of arcade facade LoRA models.

## INTRODUCTION

The historical quarters of urban environments hold significant implications in the orchestration of urban renewal initiatives. By preserving and maintaining these sectors, we perpetuate the unique cultural identity of cities, bolster cultural tourism and economic growth, and elevate urban allure and competitiveness. Nevertheless, amidst the praxis of revitalizing these historic districts, the aesthetic amelioration of architectural facades emerges as a daunting and multifaceted challenge, necessitating architects to give paramount consideration to the cultural significance and historical context of the structures. Architects must employ meticulous design and planning strategies to ensure that the rejuvenated districts persist in rendering cultural and historical contributions to the city and its communities. Moreover, the conventional architectural design process, a laborious cycle encompassing initial conceptualization, refinement, evaluation, and redesign, may involve a substantial degree of repetitious toil, especially when managing large-scale urban renewal projects.

The burgeoning technology of Artificial Intelligence Generated Content (AIGC) offers a novel paradigm for the development of revitalization schemes for historic urban quarters. Artificial intelligence generated content leverages machine learning and deep learning algorithms to facilitate architects in the effective and optimized execution of projects. Recent machine learning models have demonstrated commendable advancements in generating images of architectural exteriors. However, owing to the complex and diverse nature of facade renovation projects, the production of high-quality images remains a challenge (Q. Yu, Malaeb, and Ma 2020). This is manifest in issues of insufficient detail accuracy, limited stylistic diversity, and so forth, necessitating further intervention from the designer. By comparison, state-of-the-art stable diffusion models exhibit enhanced directionality and specificity, generating a plethora of high realistic and high-quality images in response to specific prompts swiftly (Ma 2023).

The objective of this research is to proffer a methodology for autonomously generating images of the exteriors of historic arcade buildings using stable diffusion models, with arcade architecture as a representative case, with the aim to provide architects with an inspiring and efficient workflow for revitalization designs. Stable diffusion models are employed for text-to-image generation tasks, crafting intricate images from textual descriptions. Architects, through the input of prompt and backward prompt words, can discover broader inspirations in urban renewal projects for generating arcade facades. Concurrently, we train the corresponding LoRA model (Hu et al. 2021) to ensure that architects can adjust the stylistic features of the facade images generated by the stable diffusion model, thus enhancing the controllability of the generated image content. In addition, architects can exploit the ControlNet model to further augment the controllability of images. The innovations of this paper are briefly summarized as follows: (1) We utilize diffusion models to learn the facade styles of arcade buildings in urban historic quarters. Architects can generate arcade facades by inputting professional terms, yielding inspiring reference schemes; (2) We also provide architects with an open-source LoRA model, inclusive of an arcade facade style dataset.

## RELATED WORKS

### Building Facade Generation via GAN

The application of Generative Adversarial Networks (GANs) in the field of architectural exterior generation has received substantial attention. A multitude of studies and practical applications have probed the capabilities of GANs in producing realistic and diversified architectural exteriors. Isola et al. (2017) unveiled pix2pix, a software solution that harnesses GAN algorithms for image-to-image translation tasks. This innovative approach provides an efficacious method to transmute images from one category to another, particularly when paired image data are available. Larrain, Valencia, and Yuan (2021) employed the StyleGAN algorithm to scrutinize the image database of Chilean houses constructed between the years 2010 and 2020, identifying spatial attributes suggestive of specific architectural styles inherent to Chilean housing. Sun, Zhou, and Han (2022) relied on CycleGAN to automate the generation of architectural exteriors for historic city renovation. However, issues such as inadequate detail performance, model complexity, and instability during training have been observed with GAN models in several studies (J. Zhang, Fukuda, and Yabuki 2021).

### Image Synthesis via Stable Diffusion

Beyond the aforementioned GAN models, stable diffusion models have also been deployed in the architectural realm. These models are intended for text-to-image generation tasks, crafting intricate images from textual descriptions. Ruiz et al. (2023) illustrated that large diffusion models exhibit reduced adaptability in creating architectural exteriors, with the adjustment of training and generated results often presenting challenges. Hu et al. (2021) developed a novel technique termed Low-Rank Adaptation (LoRA) of Large Language Models (LLMs), primarily designed to address the issue of fine-tuning large models. Low-rank adaptation models can freeze the weights of pre-trained models and infuse trainable rank factorization

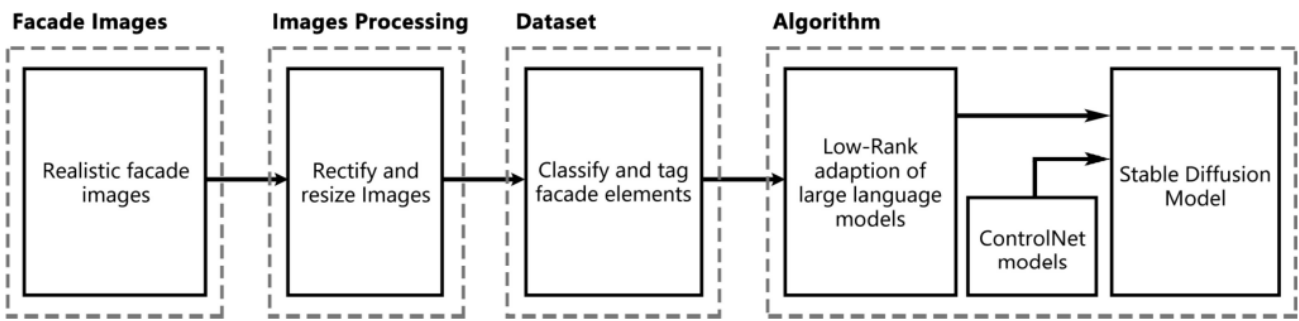

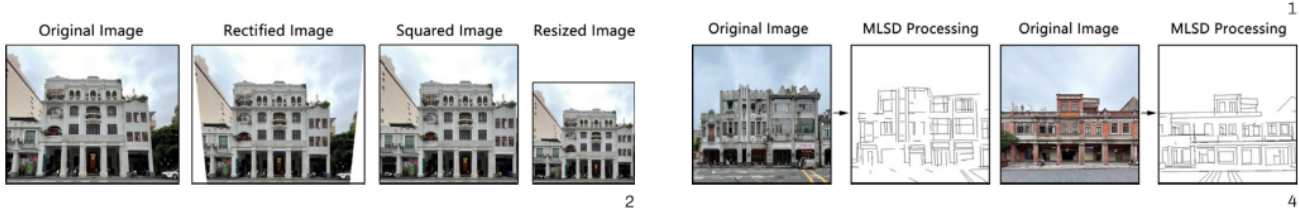

1   Methodology workflow.

2   Image processing

3   Image tags.

4   ControlNet MLSD model processing.

matrices at every layer of the transformer architecture. This significantly reduces the quantity of trainable parameters for downstream tasks, rendering it more efficient than full fine-tuning (Hu et al. 2021). Recently, the ControlNet model has been introduced to modulate the output of stable diffusion models (L. Zhang and Agrawala 2023). By incorporating additional conditions, ControlNet enables superior control over the generated images. Ma (2023) employed ControlNet to modulate the results of future generation processes, and the ControlNet's Canny Edge model can be utilized to explore different architectural facade styles and assess the impact of various weights on the generated results. Using diffusion models for generating architectural facades offers a wider range of choices for preserving the appearance of historic districts.

### Arcade Building Preservation

Arcade architecture, a distinctive type of architecture found in Lingnan, China, has formed a unique appearance through historical changes and cultural integration, becoming an integral part of architectural heritage. With urban development, the preservation of arcade historical and cultural districts faces grave challenges. Many regions have protected arcades through the establishment of cultural heritage preservation projects, for instance, the urban renewal project on Enning Road (Tan and Altrock 2016), which houses Guangzhou's most preserved and longest historical arcade district. To safeguard the historically valuable arcade buildings therein, researchers have studied and adjusted the framework of cultural preservation and planning schemes. Huang (2017) classified different styles of arcade architecture based on typology, and created several datasets of real arcade architectural facades, contributing to the digitized preservation and renewal of arcade facades. Existing research suggests that the preservation and renewal of arcade architecture still harbors significant flaws, warranting considerable efforts towards the protection and revitalization of arcade architecture.

## METHODOLOGY AND DATASET
### Methodology

Table 1: Training parameters.

| Color | LoRA model | Batch size | Epoch | Repeat | Learning rate | Optimizer type |
|---|---|---|---|---|---|---|
| | ArcadeFacadeV2.1 | 3 | 20 | 6 | 0.0001 | AdamW8bit |
| | ArcadeFacadeV2.2 | 3 | 20 | 8 | 0.0001 | AdamW8bit |
| | ArcadeFacadeV2.3 | 3 | 20 | 6 | 0.0001 | Lion |
| | ArcadeFacadeV2.4 | 2 | 20 | 6 | 0.0002 | Lion |
| | ArcadeFacadeV2.5 | 2 | 20 | 10 | 0.0002 | Lion |
| | ArcadeFacadeV2.6 | 2 | 20 | 8 | 0.0002 | Lion |

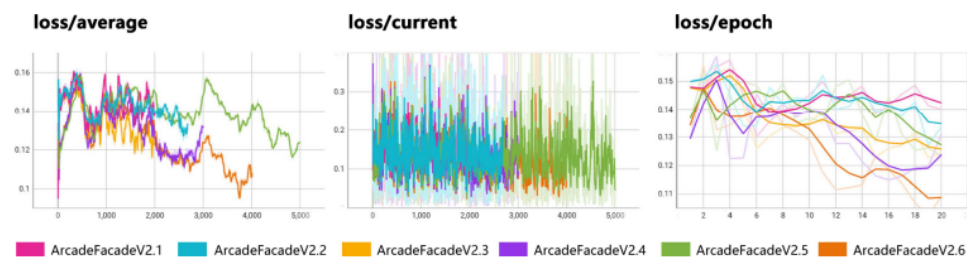

5   Training loss of arcade facade LoRA models.

This study aims to propose a method that identifies and collects images of historic architectural facades in specific urban areas, and further applies them to urban renewal transformations, ensuring the continuation of the city's historical context. The workflow, optimized through experimentation, is illustrated in Figure 1. Firstly, a series of facade images in the anticipated style are taken from historical districts, and processed. Then, elements on each facade image are distinguished and labeled with text tags, forming the training dataset for the LoRA model. Finally, based on the LoRA model and ControlNet model, architects can automatically generate facade images of specific historical styles using the Stable Diffusion model based on text or reference images. This study aims to propose a method that identifies and collects images of historic architectural facades in specific urban areas, and further applies them to urban renewal transformations, ensuring the continuation of the city's historical context. The workflow, optimized through experimentation, is illustrated in Figure 1. Firstly, a series of facade images in the anticipated style are taken from historical districts and processed. Then, elements on each facade image are distinguished and labeled with text tags, forming the training dataset for the LoRA model. Finally, based on the LoRA model and ControlNet model, architects can automatically generate facade images of specific historical styles using the Stable Diffusion model based on text or reference images.

### Image Dataset And Dataset Processing

This study chose the historical district of arcade architecture in Guangzhou city as the research object, and built an arcade facade image dataset with paired text tags. The dataset contains 50 arcade facade images, each preprocessed and assigned text tags generated by DeepDanbooru, an anime-style image tag estimation system. Figure 2 exhibits the preprocessing, including image correction, squaring, and resizing to a uniform resolution of 512x512 pixels. Based on the generated text tags, each image was assigned LoRA model trigger tags and specific text tags according to the classification of arcade facade feature elements. Specific text tags include parapet styles, balcony forms, Roman column styles, arches, line forms, and viewing angles of the arcade facades. Figure 3 shows examples of arcade facades and their paired text tags.

### Arcade Facade Generation Model

Our study heavily relies on Stable Diffusion, an advanced and complex latent diffusion model, in the tool for generating urban historical district style facades. The emergence of Stable Diffusion has brought new possibilities to our research. Stable Diffusion is a powerful latent diffusion model that can independently generate new graphics based on specific prompt words, and also recreate and draw on the basis of existing images. To ensure the accuracy, diversity, and efficiency of generated images, we further introduced the LoRA model and the ControlNet model. The joint application of these two models enables us to guide the renovation and transformation of urban historical districts more effectively and delicately, especially the generation of arcade facades.

The LoRA model is a powerful and practical pre-trained model with the ability to fine-tune large models. Even with a smaller dataset, it can train smaller models with

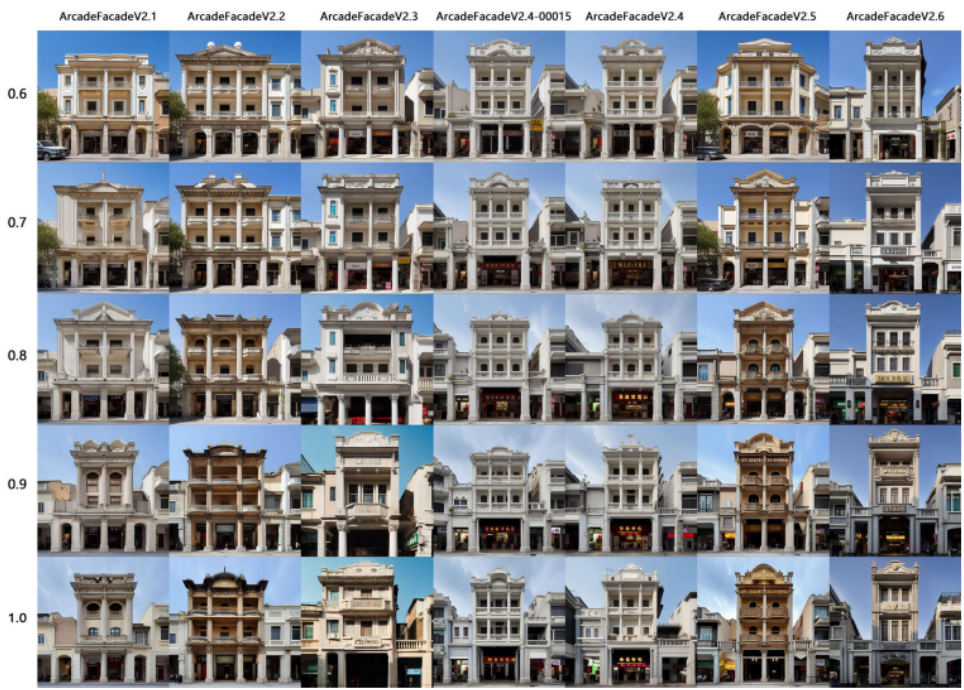

6 Training results of seven LoRA models.

fine-tuning quality comparable to full models more quickly, greatly improving our research efficiency. ControlNet is a neural network architecture that can enhance the pre-trained image diffusion model based on image prompts and control conditions provided by the user. The Mobile Line Segment Detection (MLSD) model in ControlNet serves as a preprocessor for facade images, capable of detecting and extracting straight lines from the facade images. The pre-processed images resemble architects' drafts, allowing Stable Diffusion to have more room to play. Figure 4 shows the results of image processing using the MLSD model in ControlNet.

In Conclusion, by using the LoRA model, trained on a custom arcade building facade image dataset, and the ControlNet model that takes sketches or existing images as conditional input, Stable Diffusion can accurately and faithfully generate arcade facades designed by architects using prompt words. This strategy of joint use not only makes full use of the advantages of each model, but also further improves the quality of images generated by the model, ultimately achieving the optimization and improvement of the tool for generating urban historical district style facades.

## RESULTS

### Arcade Facade LoRA Model

In our research, based on the same arcade facade image dataset and the v1-5-pruned base model, we successfully trained six LoRA models on a NVIDIA GeForce RTX 3090 Graphics Card. Table 1 provides detailed information on the training parameters of these six LoRA models. During the training process, we fine-tuned the model by adjusting key parameters, such as batch size, repeat, learning rate, and optimizer type to enhance the model's performance. We recorded the loss values of the model in real time during the training process, which served as an essential basis for evaluating the learning effect of the model and adjusting our training strategy. Figure 5 shows the loss values of the six LoRA models.

We conducted a comprehensive and detailed comparison

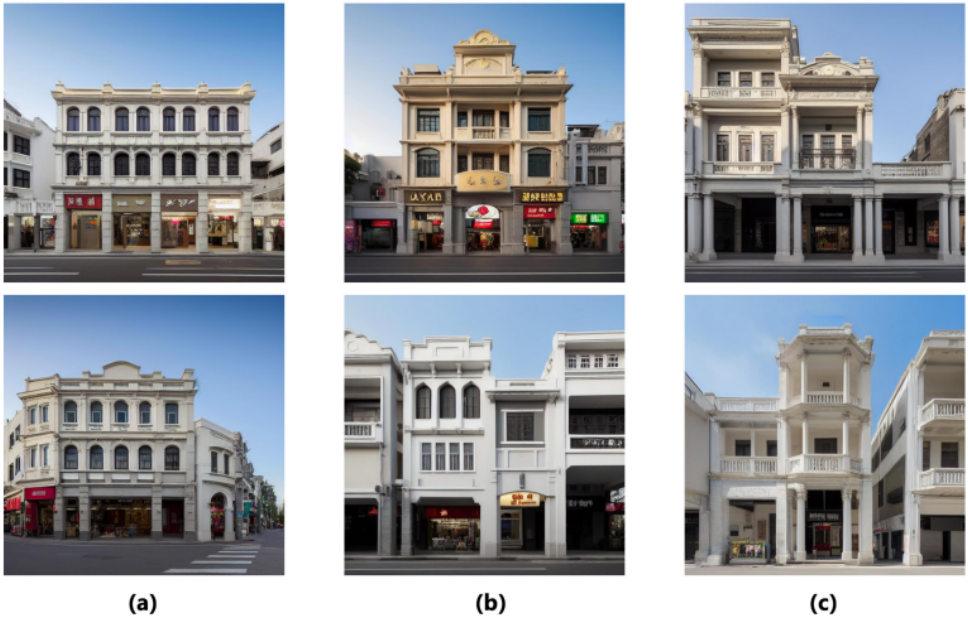

**Prompt:** real, (realistic), photo realistic, highly detailed, (masterpiece), (high quality), best quality, super detail, 4K, 8K, no humans, scenery, building, outdoors, road, shop, city, window, day, street, house, blue sky, clear, shops in hallway, arcade facade, Roman orders, Corinthian Column, undulating parapet, convex balcony, <lora:ArcadeFacadeV2.4:0.7>
**Negative prompt:** low quality, worst quality, normal quality, poor quality, soft, blurry, drawing, sketch, ugly, paintings, sketches, low resolution, tree, leaf, car, human,
**Details:** Steps: 20, Sampler: Euler a, CFG scale: 10, Seed: -1, Size: 512x512, Model hash: cc6cb27103, Model: v1-5-pruned-emaonly, Clip skip: 2, ENSD: 31337, Version: v1.2.1

(a)  (b)  (c)

7  Arcade facade generation based on prompt, (a) "street" change to "street corner", (b) "undulating parapet" change to "straight parapet", (c) "convex balcony" change to "concave balcony".

test of arcade facade generation on the six LoRA models, and the training model obtained from the 15th epoch of the fourth LoRA model, to ensure that the model we chose has the best performance. Figure 6 shows the comparison of arcade facades generated from "text-to-image" using the LoRA model in Stable Diffusion. The horizontal axis represents different LoRA models, and the vertical axis represents the weight values of the LoRA model. From the comparison, we can see that different LoRA models generate different responses to the prompt words under different weight values. Among them, ArcadeFacadeV2.4 performs well when the weight value is between 0.7 and 0.9, generating arcade building facades that fully and correctly trigger the tags in the prompt words, with the overall picture being clear and bright. Taking various factors into account, we eventually chose the excellently performing ArcadeFacadeV2.4 for fine-tuning the large model in Stable Diffusion that generates arcade facade images from prompt words. When compared with real images, a certain level of arcade building facade style extracted by the model from the dataset can be observed. These styles are not simple replications of the facade images in the dataset, but a comprehensive learning and understanding of various aspects, such as component elements, pattern engravings, material characteristics, and combination relationships, showing the intelligence and deep learning capability of our model. These results also indicate that our model can provide robust technical support for the renovation and transformation of urban historical districts.

Arcade Facade Generation via Prompt
Figure 7 clearly demonstrates the results of arcade facades generated using different text prompts, reflecting the flexibility and precision of our model. Through these results, we can see how the model flexibly responds to various prompts to generate accurate arcade facade images. In Figure 7, we use the same prompt, and change only one corresponding tag to generate two arcade facade images for comparison. For example, in part (a), changing the "street" in the prompt to "street corner" transforms the perspective of the generated arcade facade from a straight street view to a street corner view; in part (b), changing "undulating parapet" to "straight parapet" transforms the type of arcade facade parapet from undulating to flat; in part (c), changing "convex balcony" to "concave balcony" transforms the type of arcade facade balcony from a concave balcony to a convex balcony.

We use professional architectural vocabulary to describe the elements contained in the generated image, including descriptive words for picture quality, architectural types, environmental settings, spatial relationships, architectural components, and facade materials. This detailed description method can guide the model to generate images that meet the requirements more accurately. As can be seen from the results in Figure 7, the generated arcade facades conform to the spatial relationship of "upper floor, lower corridor" and can correctly trigger the specific component element tags input. The generated component elements are not only reasonably located on the arcade facade, but also demonstrate extremely innovative and diverse designs, based on the style provided in the dataset.

Arcade Facade Renewal via Prompt And ControlNet
We attempt to apply this innovative approach to the renovation and transformation of historical arcade street blocks in the city, bringing more efficient and personalized solutions to urban renewal. Using sketches drawn by architects, and on-site photos provided, we input this constraint information into ControlNet as the control conditions for the renovation and transformation of the arcade facade. The use of these conditions greatly improves the accuracy and relevance of the images generated by our model. Figure 8 shows the results of the arcade facade renovation generation based on prompt descriptions and existing photos. The results show that when we use the preprocessed results of the MLSD model in ControlNet, as control condition inputs, the generated arcade facades show controlled changes compared to before the renovation. These changes are not only limited to the building facade, but can even affect the overall appearance of the entire block, making it more in line with our design goals. These controllable changes include the addition or deletion of facade components, style changes, material replacements, and significant improvements in overall coordination. These meticulous modifications are all automatically completed by our model, greatly saving design and revision time, while also ensuring the quality of the design scheme. This approach enhances the efficacy and ingenuity of the design workflow, expedites the design process, optimizes resource utilization, and holds the potential to augment the quality of urban redevelopment. Consequently, it makes a substantial contribution to the evolution and transformation of urban landscapes.

## DISCUSSION AND CONCLUSION

In this study, we found that model parameter settings have a significant impact on training results, particularly the outstanding performance of the ArcadeFacadeV2.4 model with a weight value of 0.7 to 0.9, which demonstrates our success in model adjustment and parameter setting, thereby providing valuable reference for future model training. We noticed that the proposed workflow can accurately generate expected arcade facade images based on different text prompts, such as architectural elements, colors, materials, et cetera. It shows its precise understanding and execution ability of architectural facade elements, as well as flexible adaptation to design styles. In addition, this method has shown enormous potential when applied to the renovation of historical arcade street blocks in cities. We are able to generate arcade facades that meet the architects' requirements for facade style, size, and form, based on existing images and prompt word descriptions. For instance, we can make targeted changes to components, style changes, material replacements, and overall coordination improvements, based on specific architectural elements, historical and cultural backgrounds, and urban planning needs. This outcome showcases its controllability, accuracy, and diversity.

Our contributions can be summarized in three points: (1) The workflow we proposed not only provides a brand-new idea theoretically, but also proves through actual application that it can successfully and efficiently generate high-fidelity facade images that conform to the overall historical architectural style of the study area. This result provides valuable inspiration and choices for architects in the early design stage of urban renewal and transformation. (2) The method we proposed takes into account various restrictive factors in the process of renovating and transforming historical buildings, such as environment, historical culture, et cetera. By using current situation photos, hand-drawn sketches, and language descriptions, architects can more effectively control and guide the generation results to better achieve the design goals. (3) The workflow we proposed further drives the role transition of architects from pure designers to decision-makers and innovators. Architects' deep understanding of urban history, architectural styles, and facade composition not only drives the depth of architectural design, but also provides a strong impetus for the renewal and transformation of the city's appearance.

Although we recognize these advantages, there are also some limitations that deserve further research. For example, due to the difference between the facade image tags in the dataset processed using DeepDanbooru and reality, we need to manually input new tags. To solve this problem, we will explore new automated tag generation

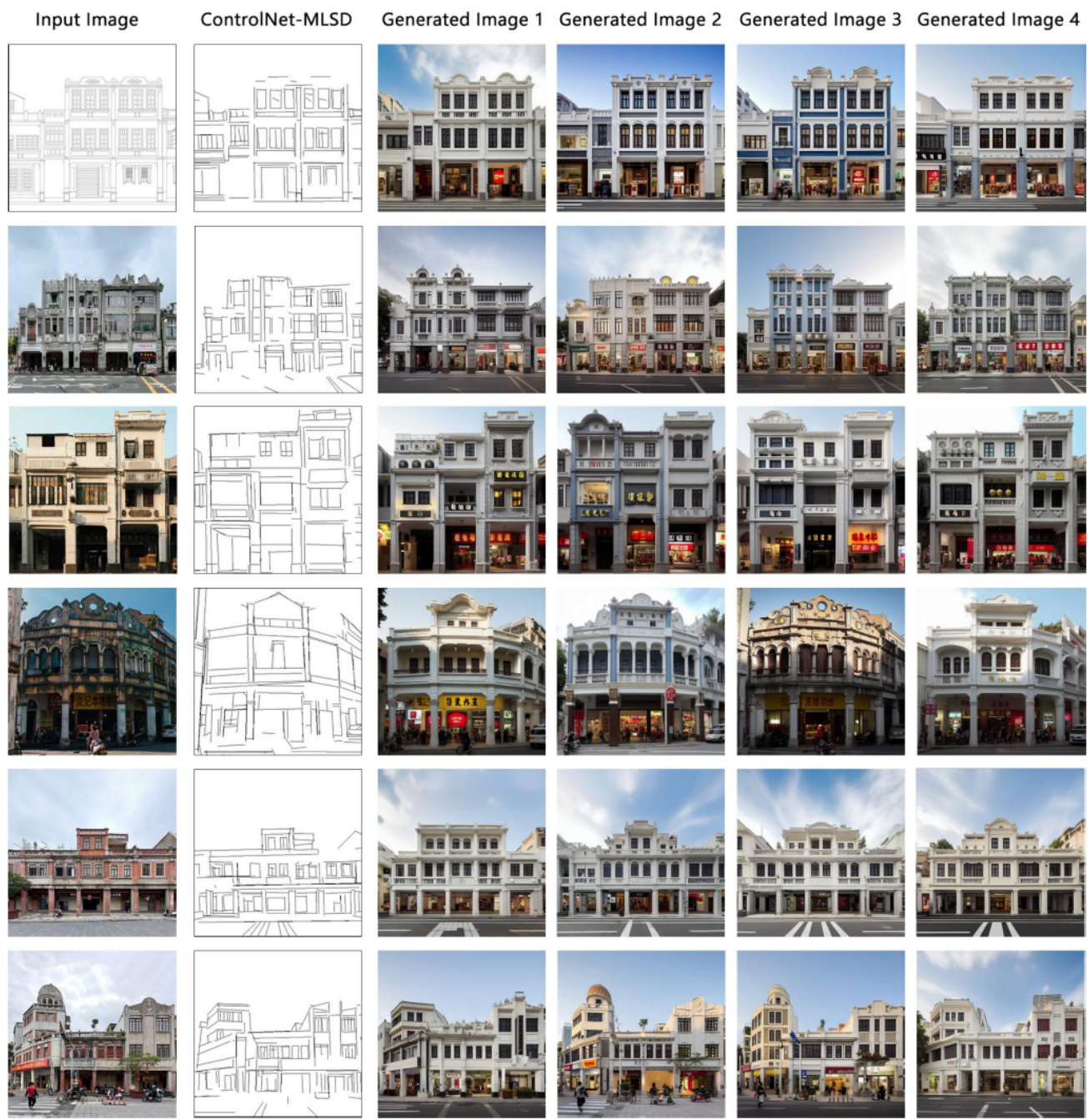

8   Arcade facade renewal based on prompt and ControlNet.

methods to improve efficiency and accuracy. In terms of the metrics for evaluating the quality of generated facade images, conventional metrics cannot reflect the perceptual realism of images. We are considering combining human evaluation to provide a more comprehensive understanding of the quality of generated images.

Simultaneously employing AIGC technology, we concurrently acknowledge the potential ethical dilemmas that may surface. As machine learning models reinterpret history, and generate these historically significant facades, it prompts us to question what it signifies when machines replicate history. Given the possibility that the images generated by these models may not accurately depict the authentic design of these edifices, we risk creating an inaccurate portrayal of history. Furthermore, the potential for historical and cultural appropriation introduces additional ethical conundrums. We must consider the social and cultural ramifications of exploiting the unique historical and cultural significance of these structures without requisite permissions. In the worst-case scenario, this could lead to the obliteration of cultural distinctiveness. As we make strides in this exciting new epoch of design and architecture, it is incumbent upon us to harbor deep reverence for historical and cultural heritage, comprehend and respect the historical and cultural backdrop of the buildings that feed into the models, enhance the transparency and openness of our models, and unequivocally indicate the potential inaccuracies in the images generated by the machine learning models.

We have demonstrated the feasibility of automatically generating new facades of urban historical block buildings through text tags and existing images. This application is expected to be utilized in promoting urban renewal and transformation. In addition, we also notice the potential of using single-image inverse rendering techniques with multi-view self-supervision (Y. Yu and Smith 2021), which will provide inspiration for this research. In future work, we will attempt to use inverse rendering techniques to generate architectural facade 2D images to 3D models, further expanding the scope and impact of our work.

## IMAGE CREDITS

All drawings and images by the authors.

**Zheyuan Kuang** is a senior Architecture student at Nanchang University in China. He possesses a profound interest in deep learning and computer graphics, as well as data-driven urban analysis and design. He has spearheaded a college-level scientific research project on digital design and intelligent construction.

**Jiaxing Zhang** began his academic journey with a Bachelor's degree in Architecture from Nanchang University in 2016, followed by a Master's degree from Southeast University in China in 2019. He earned a Ph.D. from Osaka University (Japan) School of Energy and Environment in 2022. Since November 2022, he has been a specially-appointed researcher at Osaka University's Laboratory of Environmental Design and Information Technology.

**Yiying Huang** is a junior, majoring in architecture at the School of Architecture and Design at Nanchang University, She is currently researching digital design of architecture. She has participated in research on human settlements and rural revitalization workcamp. She served as team leader.

**Yunqin Li** holds a Bachelor's degree from Nanchang University (2016), a Master's degree from Southeast University (2019), and a Ph.D from Osaka University (2022). Currently, she serves as a Visiting Researcher at Osaka University. Her research encompasses spatial auditing, measuring, perception, understanding, and interaction, backed by new data, technologies, and methods. She has a particular interest in street perception and explainable machine learning. Despite her achievements, Dr. Li remains humble in her continuous exploration of the intricate relationship between technology, space, and urban environments.